\documentclass[runningheads]{llncs}
\usepackage{graphicx}
\usepackage{subcaption}
\usepackage{algorithm,algorithmic}
\usepackage{amsmath}
\usepackage{makecell}
\usepackage{array}
\begin{document}
\title{Fake News Detection System using XLNet model with Topic Distributions: CONSTRAINT@AAAI2021 Shared Task}
\titlerunning{Fake News Detection System using XLNet model with Topic Distributions}
\author{Akansha Gautam\thanks{All authors contributed equally.}\inst{1}\orcidID{0000-0002-3172-1885}\and
Venktesh V\inst{1}\orcidID{0000-0001-5885-2175}\and
Sarah Masud\inst{1}\orcidID{0000-0003-2668-5581}}
\authorrunning{Akansha Gautam et al.}
%
\institute{Indraprastha Institute of Information Technology, Delhi \and
\email{\{akansha16221,venkteshv,sarahm\}@iiitd.ac.in}}
\maketitle              
\begin{abstract}
With the ease of access to information, and its rapid dissemination over the internet (both velocity and volume), it has become challenging to filter out truthful information from fake ones. The research community is now faced with the task of automatic detection of fake news, which carries real-world socio-political impact. One such research contribution came in the form of the Constraint@AAA12021 Shared Task on COVID19 Fake News Detection in English. In this paper, we shed light on a novel method we proposed as a part of this shared task. Our team introduced an approach to combine topical distributions from Latent Dirichlet Allocation (LDA) with contextualized representations from XLNet. We also compared our method with existing baselines to show that XLNet $+$ Topic Distributions outperforms other approaches by attaining an F1-score of $0.967$.

\end{abstract}

\section{Introduction}
With an increase in the adoption of social media as a source of news, it has become easier for miscreants to share false information with millions of users. Such activities increase during a time of crisis where some groups try to exploit the human vulnerability. One saw during COVID19 the impact of fake news\footnote{https://news.un.org/en/story/2020/04/1061592} from 5G towers to fad remedies, some even leading to physical harm. Given the volume of fake news generated on a regular basis, there is a need for automated identification of fake news to aid in moderation, as manual identification is cumbersome and time-consuming.

Fake news detection is a challenging problem because of its evolving nature and context-dependent definition of what is fake \cite{ejss}. For instance, a message shared may have a falsifiable claim but was not shared with the intent to spread misinformation. On the other hand, messages transmitted with the intent to mislead the masses may contain conspiracy theories. These messages may also include some facts that are not related to the message. While it is relatively easy for a human to identify that the facts mentioned have no relation to the claim made, it is challenging to classify news with such facts as fake automatically. It would require more training samples to induce more discriminatory power in the learned distributed representations. Automatic fake news detection has recently gained interest in the machine learning community. Several methods have been proposed for automatic fake news detection. While initial methods leverage hand-crafted features based on n-grams and psycho-linguistics \cite{perez}. Recently, rather than leveraging hand-crafted features, automatic extraction of relevant features in the form of distributed representations has become popular \cite{wang2017liar}. Various previous studies \cite{howard2018universal,eisenschlos2019multifit,sun2019fine} have shown the effect usage of Language Model Fine-tuning are an better alternative for the classification tasks than other methods.

In this paper, we propose a novel method that combines the contextualized representations from large pre-trained language models like BERT \cite{bert} or XLNet \cite{yang2019xlnet} with Topic distributions from Latent Dirichlet Allocation (LDA) \cite{10.5555/944919.944937} for the COVID19 Fake News Detection in English competition \cite{patwa2021overview}. We observed that the topic distribution provides more discriminatory power to the model. The joint representations are used for classifying the inputs as `fake' or `real'. Since the given shared task contains domain-specific language, we posit that topic distributions help provide additional signals that improve overall performance. The topic models have been previously exploited for domain adaptation \cite{hu-etal-2014-polylingual}. 

Our core technical contributions are in four areas: 
\begin{itemize}
    \item We propose a novel system for fake news detection that combined topic information and contextualized representations. (Section \ref{sec:proposed_method})
    \item We provide an extensive comparison with other states of art the neural methods and rudimentary machine learning models. (Section \ref{subsec:comparision}) 
    \item We attempt to perform error analysis both in terms of term-token counts and attention heads. (Section \ref{sec:error_analysis})
    \item We provide the source code\footnote{
Source code available at: \url{https://github.com/VenkteshV/Constraint2021}} use for modeling and error analysis along with values of hyper-parameters. (Section \ref{subsec:implementation})
\end{itemize}.

\section{Related Work}

Several researchers have already contributed by designing a novel approach to solving the problem of automatic fake news detection. A group of researchers \cite{perez} developed two datasets named \textit{Celebrity} and \textit{FakeNewsAMT} that contains equal proportions of real and fake news articles. They use linguistic properties such as n-grams, punctuation, psycho-linguistic features, readability, and Syntax to identify fake articles. They use linear SVM classifier as a baseline model to conduct several experiments such as learning curve and cross-domain analyses with a different combination of features set.

Another group of researchers \cite{ruchansky2017csi} identified the characteristics of fake news articles into three parts: (1) textual data of article (2) response of user (3) source users promoting articles. They proposed a model called CSI composed of Capture, Score, and Integrate modules. The first module uses the Recurrent Neural Network(RNN) to capture the temporal representations of articles. The second module is based on the behavior of users. The third module uses the output produced by the first two models to identify fake news articles. Some prior studies \cite{shu2017fake} have also used news content with additional information (social context information) to build a model to detect fake news. In a parallel study, \cite{long2017-fake} of fake news in China, the hybrid model assimilates the speaker profile into the LSTM. The research shows that speaker profiles help in improving the Fake News Detection model's accuracy. 

A study \cite{wang2017liar} used LIAR dataset. They proposed a model based on surface-level linguistic patterns. The baseline includes logistic regression, support vector machines, long short-term memory networks, and a convolutional neural networks model. They designed a novel, hybrid convolutional neural network to integrate metadata with text, which achieved significant fine-grained fake news detection.

A group of researchers \cite{agirrezabal2020ku} presented a robust and straightforward model for the Shared Task on profiling fake news spreaders. Their method relies on semantics, word classes, and some other simple features and then fed these features to the Random Forest model to classify fake news. 
The study \cite{liu2018early} focuses on introducing a novel method for detecting fake news on social media platforms. They used news propagation paths with both recurrent and convolutional networks to capture global and local user characteristics.  

A recent study \cite{JCS2019} presented a new set of features extracted from news content, news source, environment. It measured the prediction performance of the current approaches and features for the automatic detection of fake news. They have used several classic and state-of-the-art classifiers, including k-Nearest Neighbors, Naive Bayes, Random Forest, Support Vector Machine with RBF kernel, and XGBoost to evaluate the discriminative power of the newly created features. They measure each classifier's effectiveness with respect to the area under the ROC curve (AUC) and the Macro F1-score. Another recent study \cite{saikh2020deep} focuses on two variations of end to end deep neural architectures to identify fake news in the multi-domain platform. The first model is based on Bidirectional Gated Recurrent Unit (BiGRU) comprised of (1) Embedding Layer (2) Encoding Layer (Bi-GRU) (3) Word-level Attention (4) Multi-layer Perceptron (MLP). However, another model is based on Embedding from Language Model (ELMo) and the MLP Network.

\section{Proposed Method}
\label{sec:proposed_method}
\begin{figure*}[!ht]
\centering
\includegraphics[width=\linewidth]{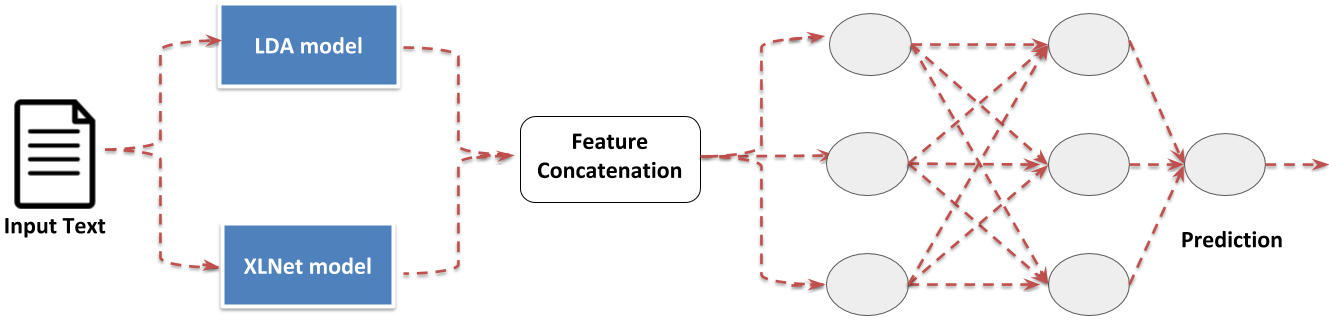}
\caption{{Proposed Model Architecture using XLNet with Topic Distributions, where contextualized representations and topic embeddings are obtained from the XLNet and LDA model, respectively. These representations are then concatenated and fed to the 2-fully connected layer followed by a Softmax Layer for the task of fake news detection.}}
\label{fig:overview}
\end{figure*}

This section describes in detail the proposed approach. The proposed neural network architecture for the fake news detection task is shown in Figure~\ref{fig:overview}. We leverage contextualized representations from XLNet and representations obtained from Latent Dirichlet Allocation (LDA) to obtain useful representations for fake news classification. The LDA is a generative probabilistic model. Each word in the document $d$ is assumed to be generated by sampling a topic from $d's$ topic distribution $\theta^d$ and then sampling a word from the distribution over words denoted by $\phi^t$ of a topic.

We leverage contextualizes representations to handle the problem of polysemy. The problem of polysemy occurs when the same word has different meanings in different contexts. The vector representations obtained through methods like word2vec are unable to disambiguate such terms and hence output the exact representations for the word irrespective of the context of their occurrence. The recent wave of pre-trained language models is based on the transformer architecture, which uses a mechanism called self-attention. The self-attention mechanism computes better representations for a word in a sentence by scoring other words in the sentence against the current word. This helps determine the amount of focus placed on different input sequence words when computing the present word's vector representation. The pre-trained language model BERT \cite{bert} was built using the transformer and provided useful contextualized representations for many downstream tasks. However, there were several drawbacks to BERT. During training, the BERT model predicts the masked tokens in parallel. This may result in wrong predictions as the value of one of the masked tokens may depend on another masked token. For instance, for the sentence ``I went to the [MASK] to get [MASK]". Here, the words ``hospital" and ``vaccinated" for the first and second masks are more probable than the words ``hospital" and ``coffee". However, there are many possible combinations when the BERT model predicts the tokens in parallel, resulting in an incorrect sentence. The XLNet model \cite{yang2019xlnet} helps overcome certain drawbacks of BERT by introducing the \emph{permutation language modeling} and by using transformer-XL architecture \cite{dai-etal-2019-transformer} as the backbone.  The transformer-XL architecture introduces the recurrence mechanism at the segment level to the transformer architecture. It accomplished this by caching the hidden states generated from the previous segment and uses them as keys and values when processing the next segment. The permutation language modeling method predicts one token at a time given the preceding context like a traditional language model. However, it predicts the tokens at random order rather than the sequential one. Hence, the permutation language modeling does not need the [MASK] tokens and does not have independent parallel predictions observed in BERT.

In the proposed method, the news article (denoted as $a_{i}$) is passed through XLNet model to obtain contextualized representations (denoted as $CE(\cdot)$). The LDA model is trained on the provided training set and is then leveraged to compute the document-topic embeddings (denoted as $TE(\cdot)$) for a given input. The training of LDA is done only once and hence does not add to the inference time. The document-topic distributions can be pre-computed for the entire training set. This can be accomplished easily in our architecture as the computation of the document-topic distributions is decoupled from the XLNet forward pass. The final input representation can be obtained by combining the input's topic distribution with the contextualized embeddings of the sentence. We denote the final input representation as $IE$, as shown below:

\begin{equation}
       IE(a_{i}) = \big[[CE(t), TE(t)]\big| t\in a_i\big]
\end{equation}

The concatenated feature representation is passed through $2$-fully connected layers followed by a Softmax Layer to output the prediction $y_{i}$ for classification of news articles.

\begin{equation}
    y_{i} = Softmax(IE(a_{i}))
\end{equation}

We perform extensive experiments by varying the model architecture. The dataset, experiments conducted, and the baselines are discussed in the following section. 

\section{Dataset Description}

We use the COVID-19 Fake News Dataset given by \cite{patwa2020fighting}. It is a manually annotated dataset of 10,700 social media posts and articles of real and fake news based on topics related to COVID-19. Fake news articles are collected from several fact-checking websites and tools, whereas real news articles are collected from Twitter using verified Twitter handles. Table~\ref{tab:examples_dataset} depicts examples of fake and real articles from the COVID-19 Fake News Dataset.

\begin{table*}[ht]
    \centering
    \begin{tabular}{|p{2 cm}|p{9 cm}|}
         \hline 
         \textbf{Label} & \textbf{Text} \\ \hline
         
         Fake & No Nobel Prize laureate Tasuku Honjo didn't say the coronavirus is ``not natural" as a post on Facebook claims. In fact Professor Honjo said he's ``greatly saddened" his name was used to spread misinformation. This and more in the latest \#CoronaCheck: https://t.co/rLcTuIcIHO https://t.co/WdoocCiXFu \\ \hline
         
         Real & We launched the \#COVID19 Solidarity Response Fund which has so far mobilized \$225+M from more than 563000 individuals companies \&amp; philanthropies. In addition we mobilized \$1+ billion from Member States \&amp; other generous to support countries-@DrTedros https://t.co/xgPkPdvn0r \\ \hline
         
         
        
    \end{tabular}
    \caption{Examples of fake and real news articles from the dataset}
    \label{tab:examples_dataset}
\end{table*}

The dataset is split into 6420 samples in the train set and test and validation sets with 2140 samples. Table~\ref{tab:dataset_split} shows the distribution of data across $2$ different classes. It suggests that data is class-wise balanced and class distribution is similar across Train, Validation, and Test Split.   

\begin{table}[]
    \centering
    \begin{tabular}{|p{2 cm}|p{2 cm}|p{2 cm}|p{2 cm}|}
        \hline \textbf{Split} & \textbf{Real} & \textbf{Fake} & \textbf{Total}  \\ \hline
        Train & 3360 & 3060 & 6420 \\ \hline
        Validation & 1120 & 1020 & 2140 \\ \hline
        Test & 1120 & 1020 & 2140 \\ \hline
        \textbf{Total} & 5600 & 5100 & 10700 \\ \hline
    \end{tabular}
    \caption{Distribution of dataset across 2 different classes, Real and Fake}
    \label{tab:dataset_split}
\end{table}

\section{Experiments}

The proposed approach was implemented using PyTorch and with an NVIDIA Tesla K80 GPU. We use \emph{Transformers} library\footnote{\url{https://github.com/huggingface/transformers}} maintained by the researchers and engineers at Hugging Face \cite{wolf2019huggingface} which provides PyTorch interface for XLNet.
\emph{Transformers} library supports Transformer-based architectures such as BERT, RoBERTa, DistilBERT, XLNet \cite{yang2019xlnet} and facilitates the distribution of pre-trained models.

\subsection{Implementation}
\label{subsec:implementation}
The pre-processing of data involves in our approach is inspired from various sources \cite{patwa2020fighting}, \cite{kumar2020nutcracker}. We pre-processed the data by removing emojis, stopwords, special characters, hashtag symbols, usernames, links, and lowercasing the texts. We use \verb|xlnet-base-cased| model to conduct our experiment. To provide input to the XLNet model, we first split our text into tokens and mapped these tokens to their index in the tokenizer vocabulary. For each tokenized input text, we construct the following:

\begin{itemize}
    \item \textbf{input ids}: a sequence of integers identifying each input token to its index number in the XLNet tokenizer vocabulary
    \item \textbf{attention mask}: a sequence of $1$s and $0$s, with $1$s for all input tokens and $0$s for all padding tokens
    \item \textbf{topic embeddings}: a sequence of probabilities signifies the likelihood of a word in conjunction with a given topic using LDA model
    \item \textbf{labels}: a single value of $1$ or $0$. In our task, $1$ means ``Real News,” and $0$ means ``Fake News.”
\end{itemize}

The model is fine-tuned for $15$ epochs with a learning rate of $2e^{-5}$ and an epsilon value of $1e^{-8}$.

\subsection{Comparison with other methods}
\label{subsec:comparision}
We compare our results with the baseline \cite{patwa2020fighting} method and our other experimented methods. The explanation about our other methods are mentioned as follows:

\begin{itemize}
    \item \textbf{USE $+$ SVM}: We first adopt a ML-based approach. Instead of TF-IDF features, we obtain contextualized representations of the input using Universal sentence encoder (USE)\footnote{\url{https://tfhub.dev/google/universal-sentence-encoder-large/3}}. We then fed the input representations to an SVM model.
    \item \textbf{BERT with Topic Distributions}: In this approach, we combine the document-topic distributions from LDA with contextualized representations from BERT. The model was fine-tuned for $10$ epochs (with early stopping) with the ADAM optimizer, with a learning rate of $2e^{-5}$ and epsilon is set to $1e^{-8}$.
    \item \textbf{XLNet}: Here, we fine-tune the pre-trained XLNet model on the given input. This model was fine-tuned for 15 epochs with ADAM optimizer using the learning rate of $2e^{-5}$, and epsilon is set to $1e^{-8}$.
    \item \textbf{Ensemble Approach: BERT and BERT $+$ topic}: Here, we combine the predictions of the BERT and BERT $+$ topic models. This provides an increase in performance on the validation set. However, this variant does not outperform the proposed XLNet with the Topic Distributions model on the test set.
\end{itemize}

Table~\ref{tab:comparison_results} shows the performance of baseline, experimented, and final proposed method using several evaluation metrics such as Precision, Recall, and weighted F1-score on the Test set. It suggests that our proposed method outperforms the baseline and other models by achieving an F1-score of 0.967. 

\begin{table}[ht]
\centering
\caption{Performance comparison of proposed method with baseline and other variants on Test set}
\begin{tabular}{|p{7.75 cm}|p{1.5cm}|p{1.5cm}|p{1.5cm}|} 
\hline
 \textbf{Method} & \textbf{Precision} & \textbf{Recall} & \textbf{F1-score} \\ \hline
 Baseline method \cite{patwa2020fighting} & 0.935 & 0.935 & 0.935 \\
 USE $+$ SVM & 0.92 & 0.92 & 0.92 \\
 BERT with Topic Distributions & 0.949 & 0.948 & 0.948 \\
 XLNet & 0.949 & 0.948 & 0.948 \\
 Ensemble Approach: BERT and BERT $+$ topic & 0.966 &0.966 &0.966 \\
 XLNet with Topic Distributions (Proposed method) & \bf0.968 & \bf0.967 & \bf0.967 \\
\hline
\end{tabular} \label{tab:comparison_results}
\end{table}

\subsection{Results and Discussion}
The results of the comparison of the proposed method with baselines and several variants of the proposed method are shown in Table \ref{tab:comparison_results}. From the table, it is evident that including topical information enhances the performance as BERT+topic outperforms the baseline methods and is similar to the performance of XLNet. Also, XLNet with Topic Distributions outperforms all methods. We also observe that the difference in F1 scores between the ensemble approach and XLNet with Topic Distributions is not statistically significant. The above results support the hypothesis that topic distributions help in domain adaptation enhancing performance. The topic distributions are pre-computed and hence can be indexed, making our method efficient for inference.

\section{Error analysis}

\label{sec:error_analysis}
\begin{figure}[!h]
    \centering
    \includegraphics[width=0.4\linewidth]{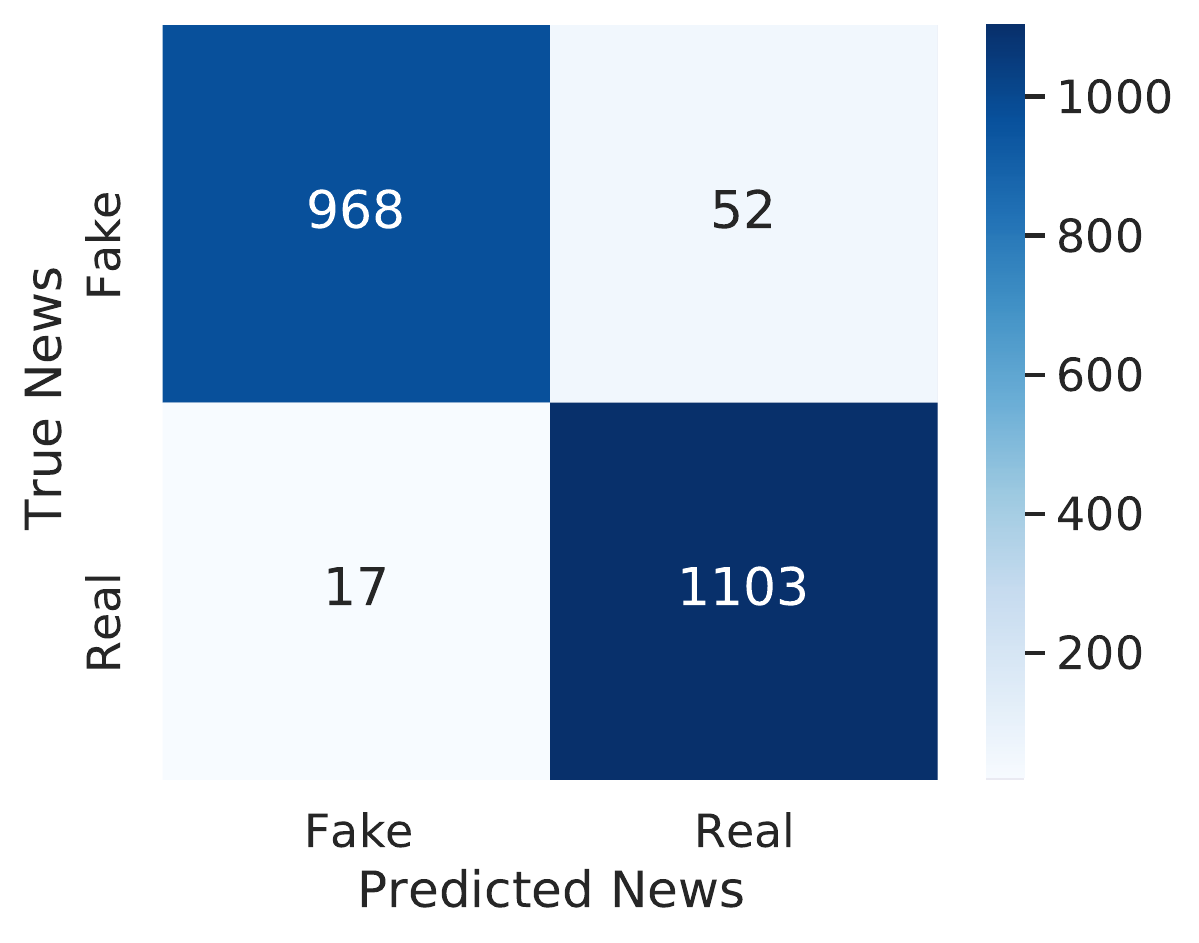}
    \caption{Confusion Matrix of proposed method on Test Set}
    \label{fig:confusion_matrix}
\end{figure}
Based on the Figure~\ref{fig:confusion_matrix}, we see that there are a total of $69$ misclassified samples. Let us look at a few of these misclassified test samples based on how the sample keywords are distributed across the fake and real classes in the training + validation set combined.

 \begin{figure}
        \includegraphics[width=0.5\columnwidth]{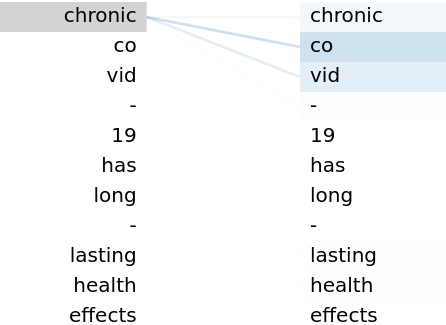}
        \includegraphics[width=0.45\columnwidth]{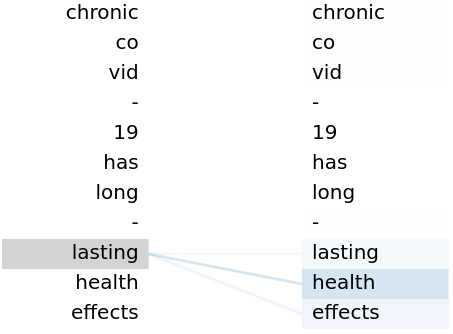}
         \caption{Attention weights for the terms ``chronic" and ``lasting" for attention head $7$ at layer $7$ of fine-tuned XLNet. Input is Example 6.}
        \label{fig:attention}
\end{figure}

\begin{table}[!ht]
\caption{Word token occurrence of some keyword in the above examples. The count is based on the combined training and validation set for the two classes.\newline}

\begin{subtable}{0.45\textwidth}
    \centering
    \scalebox{0.825}{
    \begin{tabular}{|c|c|c|}
        \hline
        \textbf{Keyword}& \textbf{Count Fake Class} & \textbf{Count Real Class}\\\hline
         people & $358$ & $581$\\
         hospital & $212$ & $141$\\
         covid-19 & $1194$ & $880$\\\hline
    \end{tabular}}
    \caption{Keyword occurrence of most contributing words in Example 1.}
    \label{tab:example_1}
\end{subtable}
\hspace{2.4em}
\begin{subtable}{0.55\textwidth}
    \scalebox{0.825}{
    \begin{tabular}{|p{1.6cm}|c|p{2cm}|}
        \hline
        \textbf{Keyword}& \textbf{Count Fake Class} & \textbf{Count Real Class}\\\hline
         claim & $139$ & $1$\\
         india's & $11$ & $59$\\
         tata & $5$ & $1$\\
         survival & $6$ & $1$\\\hline
    \end{tabular}}
    \caption{Keyword occurrence of most contributing words in Example 2.}
    \label{tab:example_2}
\end{subtable}
\begin{subtable}{0.45\textwidth}
    \centering
    \scalebox{0.825}{
    \begin{tabular}{|c|c|c|}
        \hline
        \textbf{Keyword} & \textbf{Count Fake Class} & \textbf{Count Real Class}\\\hline
         positive & $128$ & $212$\\
         coronavirus & $1590$ & $371$\\
         cases & $194$ & $2003$ \\
         ventilator & $8$ & $10$\\
         doctors & $-$ & $-$ \\\hline
    \end{tabular}}
    \caption{Keyword occurrence of most contributing words in Example 3. }
    \label{tab:example_3}
\end{subtable}
\hspace{1em}
\begin{subtable}{0.45\textwidth}
    \centering
    \scalebox{0.825}{
    \begin{tabular}{|c|c|p{2cm}|}
        \hline
        \textbf{Keyword} & \textbf{Count Fake Class} & \textbf{Count Real Class}\\\hline
         millions & $22$ & $6$\\
         different & $22$ & $39$\\
         rashes & $-$ & $-$ \\
         covid19 & $255$ & $1545$\\
         symptom & $3$ & $11$ \\\hline
    \end{tabular}}
    \caption{Keyword occurrence of most contributing words in Example 4. }
    \label{tab:example_4}
\end{subtable}
\begin{subtable}{1\textwidth}
    \centering
    \scalebox{0.825}{
    \begin{tabular}{|c|c|c|}
        \hline
        \textbf{Keyword} & \textbf{Count Fake Class} & \textbf{Count Real Class}\\\hline
         people & $358$ & $581$\\
         low & $15$ & $83$\\
         risk & $25$ & $183$ \\
         negative & $16$ & $80$ \\
         test & $97$ & $222$ \\
         results & $17$ & $67$ \\ \hline
    \end{tabular}}
    \caption{Keyword occurrence of most contributing words in Example 5.}
    \label{tab:example_5}
\end{subtable}
\begin{subtable}{1\textwidth}
    \centering
    \scalebox{0.825}{
    \begin{tabular}{|c|c|c|}
        \hline
        \textbf{Keyword} & \textbf{Count Fake Class} & \textbf{Count Real Class}\\\hline
         chronic & $3$ & $11$\\
        covid-19 & $1194$ & $880$\\
        health & $153$ & $370$ \\
         effects & $12$ & $12$ \\ \hline
    \end{tabular}}
    \caption{Keyword occurrence of most contributing words in Example 6.}
    \label{tab:example_6}
\end{subtable}

\end{table}

\begin{itemize}
    \item \textbf{EXAMPLE 1 (Test ID 351, Real Classified as Fake):} \textit{today there are 10 people in hospital who have covid-19 three people are in auckland city hospital four people in middlemore two people in north shore hospital and one person in waikato hospital he new person in auckland city hospital is linked to the community cluster}. As we observe from Table \ref{tab:example_1} that the combined negative impact of terms ``covid-19" and ``hospital" is much greater than the positive impact of the term ``people", which could explain why the prediction skews towards the ``Fake" class instead of its actual ``Real" class.
    \item \textbf{EXAMPLE 2 (Test ID 186, Real Classified as Fake):}\textit{the claim stated that india's top business conglomerate tata group chairman ratan tata said it's not time to think of profits but to think of survival}. Similar to previous example we observe (Table \ref{tab:example_2}) that the negative impact of term ``claim" is much greater than the positive impact of the word ``india's", which again causes the prediction to skew towards the ``Fake" class instead.
    \item \textbf{EXAMPLE 3 (Test ID 1145, Fake Classified as Real):} \textit{there are 59 positive coronavirus cases in nagpur along with three doctors, one of whom is on ventilator}. As we see from Table \ref{tab:example_3}, the positive impact of the terms ``positive", ``cases" and ``ventilator", outweight the negative impact of the term ``coronavirus". Now, had XLNet attention given more weightage to the negative term ``coronavirus", the predictions would have been on point, but that does not seem to be happening for this example.
    \item \textbf{EXAMPLE 4 (Test ID 468, Fake Classified as Real):}\textit{ millions of app users' send in 3900 photo's of 8 different type of rashes, so now they're a covid19 symptom www}. As we observe from Table \ref{tab:example_4}, the minor negative impact of the term ``million" is matched by the minor positive impact of the terms ``different" and ``symptoms". Meanwhile the seemingly important keyword ``rashes" is not observed at all in any of the training samples. It is however, the highly positive impact of the term ``covid19" that skews the prediction in favour of class ``Real" instead of ``Fake".
    \item \textbf{EXAMPLE 5 (Test ID 1147, Fake Classified as Real):}\textit{these people have been assessed as presenting a very low risk due to the nature of their exemption adherence to their required protocols and the negative test results of people associated with their bubble}. We see that unfortunately all keywords are contributing positively, giving way to the the prediction being ``Real" rather than ``Fake". 
    \item \textbf{EXAMPLE 6 (Test ID 663, Real Classified as Fake:)}\textit{chronic covid-19 has long-lasting health effects}. We see here that the while the while combine impact of the terms ``covid-19" and ``health" is tilted towards positive, the predicted output is ``Fake". Since, this result cannot be directly explained in terms of term count, we dig deeper and found that the overall attention given to the term ``covid" is higher than that of the term ``health". For 7th attention head, of the 7th layer (\ref{fig:attention}), un-normalised attention weight for term ``covid" is around $\approx1.0$, while that of ``health" and ``effects" combined lags at $\approx0.3$. This difference in attention weights and the skewed class-wise count have the combined affect of shifting the predicted distribution towards ``Fake".
\end{itemize}

Some of the techniques that can help reduce this bias towards count could be inclusion of theme specific stop words (common terms like doctors, tests which are related to Covid-19), weighing token weights in XLNet by tf-idf based techniques (give importantce of rare words in each class), manual mapping of abbreviations and similar terms to consolidate their impact (``covid19, covid-19,coronavirus" all point to same entity).

\section{Conclusion}
This paper has proposed a fake news detection system that exploits transfer learning with the LDA model. We used the XLNet model with Topic Distributions derived from the LDA model. Our proposed approach provides a gain in performance when compared to other neural models. We attribute the gain to the topic distributions, which provide more discriminatory power. In the future, we aim to leverage BERTweet \cite{nguyen2020bertweet}, which has been trained on a corpus of 850M English tweets. The tweets are different from traditional text from Wikipedia in terms of sequence length and frequent use of informal language. Hence, a language model pre-trained on a large corpus of tweets would help increase the performance when fine-tuned for domain specific tasks.

\bibliographystyle{splncs04}
\bibliography{aaai}

\end{document}